\def\year{2020}\relax

\documentclass[letterpaper]{article} % DO NOT CHANGE THIS
\usepackage{aaai20}  % DO NOT CHANGE THIS
\usepackage{svg}
\usepackage{times}  % DO NOT CHANGE THIS
\usepackage{helvet} % DO NOT CHANGE THIS
\usepackage{courier}  % DO NOT CHANGE THIS
\usepackage[hyphens]{url}  % DO NOT CHANGE THIS
\usepackage{graphicx} % DO NOT CHANGE THIS
\usepackage{subcaption}
\usepackage{multirow}
\usepackage{dirtytalk}
\usepackage{amsmath}
\usepackage{caption} % DO NOT CHANGE THIS AND DO NOT ADD ANY OPTIONS TO IT
\usepackage{todonotes}

\nocopyright
\title{Understood in Translation: Transformers for Domain Understanding}
\author {
    Dimitrios Christofidellis,\textsuperscript{\rm 1, \rm 2}
    Matteo Manica,\textsuperscript{\rm 1}
    Leonidas Georgopoulos,\textsuperscript{\rm 1}
    Hans Vandierendonck \textsuperscript{\rm 2}\\

\textsuperscript{\rm 1}IBM Research Europe \\ 
\textsuperscript{\rm 2} Queen’s University Belfast \\
dic@zurich.ibm.com, tte@zurich.ibm.com, leg@zurich.ibm.com,  h.vandierendonck@qub.ac.uk
}

\begin{document}

\maketitle

\begin{abstract}
Knowledge acquisition is the essential first step of any Knowledge Graph (KG) application. This knowledge can be extracted from a given corpus (KG generation process) or specified from an existing KG (KG specification process). Focusing on domain specific solutions, knowledge acquisition is a labor intensive task usually orchestrated and supervised by subject matter experts. Specifically, the domain of interest is usually manually defined and then the needed generation or extraction tools are utilized to produce the KG. Herein, we propose a supervised machine learning method, based on Transformers, for domain definition of a corpus. We argue why such automated definition of the domain's structure is beneficial both in terms of  construction time and quality of the generated graph. The proposed method is extensively validated on three public datasets (WebNLG, NYT and DocRED) by comparing it with two reference methods based on CNNs and RNNs models. The evaluation shows the efficiency of our model in this task. Focusing on scientific document understanding, we present a new health domain dataset based on publications extracted from PubMed and we successfully utilize our method on this. Lastly, we demonstrate how this work lays the foundation for fully automated and unsupervised KG generation.
\end{abstract}

\section{Introduction}

\begin{figure*}
\centering

\begin{subfigure}[b]{.38\linewidth}
\includegraphics[width=0.8\linewidth]{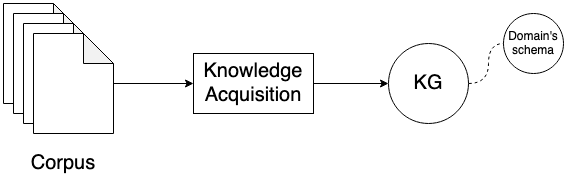}
\caption{Bottom-up pipeline.}\label{fig:general}
\end{subfigure}
\begin{subfigure}[b]{.38\linewidth}
\includegraphics[width=0.8\linewidth]{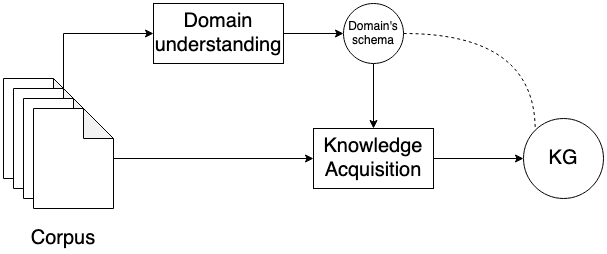}
\caption{Top-down pipeline.}\label{fig:domain}
\end{subfigure}
\caption{Typical pipelines of bottom-up and top-down KG generation.}
\label{fig:archs}
\end{figure*}

Knowledge Graphs (KGs) are among the most popular data management paradigms and their application is widespread across different fields, e.g., recommendation systems, question-answering tools and knowledge discovery applications. This is due to the fact that KGs share simultaneously several advantages of databases (information retrieval via structured queries), graphs (representing loosely or irregularly structured data) and knowledge bases (representing semantic relationship among the data). KG research can be divided in two main streams \cite{kg_survey}: knowledge representation learning, which investigates the representation of KG into vector representations (KG embeddings), and knowledge acquisition,  which considers the KG generation process. The latter being a fundamental aspect since a malformed graph will not be able to serve reliably any kind of downstream task.

\par
The knowledge acquisition process is either referred to KG construction, where the KG is built from scratch using a specific corpus, or KG specification, where a subgraph of interest is extracted from an existing KG. In both cases, the acquisition process can follow a bottom-up or top-down approach \cite{zhao2018architecture}. In a bottom-up approach, all the entities and their connections are extracted as a first step of the process. Then, the underlying hierarchy and structure of the domain can be inferred from the entities and their connections. Conversely, a top-down approach starts with the definition of the domain's schema that is then used to guide the extraction of the needed entities and connections. For general KG generation, a bottom-up approach is usually preferred as we typically wish to include all entities and relations that we can extract from the given corpus. Contrarily, a top-down approach better suits a domain-specific KG generation or KG specification, where entities and relations are strongly linked to the domain of interest. The structure of typical bottom-up and top-down pipelines, focusing on the case of KG generation, are presented in figures \ref{fig:general} and \ref{fig:domain} respectively.

 \par
 Herein, we focus on domain-specific, i.e., top-down, acquisition for two main reasons. Firstly, the acquisition process can be faster and more accurate in this way. By specifying the schema of the domain of interest, then we only need to select the proper and needed tools (i.e. pretrained models) for the actual entity and relation extraction. Secondly, such approach minimizes the presence of irrelevant data and restricts queries and graph operations to a carefully tailored KG. This generally improves the accuracy of KG applications \cite{harnessing}. Furthermore, the graph's size is significantly reduced by excluding irrelevant content. Thus, execution time of queries can be reduced by more than one order of magnitude \cite{harnessing}.

 \par
The domain definition is usually performed by subject matter experts. Yet, knowledge acquisition by expert curation can be extremely slow as the process is essentially manual. Moreover, human error may affect the data quality and lead to malformed KGs. In this work, we propose to overcome these issues by introducing an automated machine learning-based approach to understand the domain of a collection of text snippets. Specifically, given sample input texts, we infer the schema of the domain to which they belong. This task can be incorporated into both domain-specific KG generation and KG specification process, where the domain definition is the essential first step. For the KG generation, the input texts can be samples from the corpus of interest, while for the KG specification, these text snippets can express possible questions that need to be answered from the specified KG. We introduce a seq2seq-based model relying on transformer architecture to infer the relation types characterizing the domain of interest. Such model lets us to define the domain's schema including all the needed entity and relation types. The model can be trained using any available previous schema (i.e., schema of a general KG like DBpedia) and respective text examples for each possible relation type. We show that our proposed model outperforms other baseline approaches, it can be successfully utilized for scientific documents and it has interesting potential extension in the field of automated KG generation.

\section{Related work}

At the best of our knowledge, our method is the first attempt to introduce a supervised machine learning based domain understanding tool that can be incorporated into domain-specific KG generation and specification pipelines. Currently, the main research interest related to KG generation workflows is associated with attempts to improve the named entity recognition (NER) and the relation extraction tasks or provide end-to-end pipelines for general or domain-specific KG generation \cite{kg_survey}. The majority of such work focuses on the actual generation step and rely solely on manual identification of the domain definition \cite{luan2018multi,manica2019information,covid_kg}. 

As it concerns the KG specification field, the subgraph extraction is usually based on graph traversals or more sophisticated heuristics techniques and some providing initial entities or entity types \cite{harnessing}. Such approaches are effective, yet a significant engineering effort is required to tune the heuristics for each different case. Let alone, the crucial task of proper selection of the initial entities or entity types is mainly performed manually. 

The relation extraction task is also related to our work. It aims at the extraction of triplets of the form of (subject, relation, object) from the texts. The neural network based methods, such as \citeauthor{nguyen2015relation,zhou2016attention,zhang2017position}, dominate the field. These methods are CNN \cite{zeng-etal-2014-relation,nguyen2015relation} or LSTM \cite{zhou2016attention,zhang2017position} models, which attempt to identify relations in a text given its content and information about the position of entities in it. The positional information of the entities is typically extracted in a previous step of KG generation using NER methods \cite{nadeau2007survey}. Lately, there is a high interest of methods that can combine the NER and relation extraction tasks into a single model \cite{zheng2017joint,zeng2018extracting,fu2019graphrel}. 

While our work is linked to relation extraction it has two major differences. Firstly, we focus on the relation type and the entity types that compose a relation rather than the actual triplet. Secondly, the training process differs and requires coarser annotations. We solely provide texts and the respective existing sequence of relation types. Contrarily in a typical relation extraction training process, information about the position of the entities in the text is also needed. Here, we propose to improve knowledge acquisition by performing a data-driven domain definition providing an approach that is currently unexplored in KG research.

\begin{figure*}
    \begin{center}
    \includegraphics[scale=0.25]{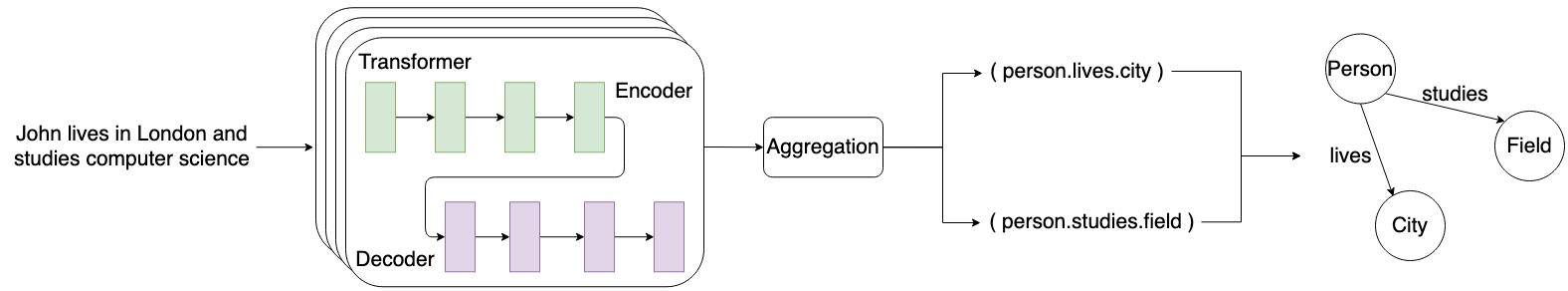}
    \end{center}
    \caption{Architecture of our utilized Transformer model for domain understanding.}
    \label{fig:arch}
\end{figure*}

\section{Seq2seq-based model for domain understanding}

The domain understanding task attempts to uncover the structured knowledge underlying a dataset. In order to depict this structure we can leverage the so called domain's metagraph. A domain's metagraph is a graph that has as vertices all the entity types and as edges all their connections/relations in the context of this domain. The generation of such a metagraph entails obtaining all the entity types and their relations. Assuming that each of entity types that are presented in the domain has at least one interaction with another entity type, the metagraph  of this domain can be produced by inferring all the possible relation types as all the entity types are included in at least one of them. Thus, our approach aims to build an accurate model to detect a domain's relation types, and leverages this model to extract those relations from a given corpus. Aggregating all extracted relations yields the domain's metagraph.

\subsection{Seq2seq model for domain's relation types extraction}

Sequence to sequence models (seq2seq) \cite{bahdanau2014neural,cho2014learning,sutskever2014sequence,jozefowicz2016exploring} attempt to learn the mapping from an input X to its corresponding target Y where both of them are represented by sequences. To achieve this they follow an encoder-decoder based approach. Encoders and decoders can be recurrent neural networks \cite{cho2014learning} or convolutional based neural networks \cite{gehring2017convolutional}. In addition, an attention mechanism can also be incorporated into the encoder \cite{bahdanau2014neural,luong2015effective} for further boosting of the model's performance.
Lately, Transformer architectures \cite{attention,bert,roberta,gpt}, a family of models whose components are entirely made up of attention layers, linear layers and batch normalization layers, have established themselves as the state of the art for sequence modeling, outperforming the typically recurrent based components. Seq2seq models have been successfully utilized for various tasks such as neural machine translation \cite{bahdanau2014neural} and natural language generation \cite{pust2015parsing}. Recently, their scope has also been extended beyond language processing in fields such as chemical reaction prediction \cite{MolecularTransformer}.

\par
We consider the domain's relation type extraction task as a specific version of machine translation from the language of the corpus to the \say{relation} language that includes all the different relations between the entity types of the domain. A relation type R which connects the entity type i to j is represented as \say{i.R.j} in the \say{relation} language. In the case of undirected connections, \say{i.R.j} is the same as \say{j.R.i} and for simplicity we can discard one of them.

\par
Seq2seq  models have been designed to address tasks where both the input and the output sequences are ordered. In our case the target \say{relation} language does not have any defined ordering as per definition the edges of a graph do not have any ordering. In theory the order does not matter, yet in practice unordered sequences will lead to slower convergence of the model and requirements of more training data to achieve our goal \cite{order}. To overcome this issue, we propose a specific ordering of the \say{relation} language influenced from the semantic context that the majority of the text snippets hold.

\par
According to \cite{zeng2018extracting}, in the context of relation extraction, text snippets can be divided into three types: \texttt{Normal}, \texttt{EntityPairOverlap} and \texttt{SingleEntityOverlap}.
A text snippet is categorized as \texttt{Normal} if none of its triplets have overlapping entities. If some of its triplets express a relation on the same pair of entities, then it belongs to the \texttt{EntityPairOverlap} category and if some of its triplets have one entity in common but no overlapped pairs, then it belongs to \texttt{SingleEntityOverlap} class. %Table \ref{table:1} presents one example for each category.
These three categories are also relevant in the metagraph case, even if we are working with entity types and relation types rather than the actual entities and their relations.

\begin{table*}[]
\centering
\resizebox{0.8\textwidth}{!}{
\begin{tabular}{|c|c|c|c|}
\hline
Dataset   & \# instances & size of \say{relation} language & mean \#relation types per instance \\ \hline
WebNLG    & 23794        & 70                              & 2.15                               \\ \hline
NYT       & 70029        & 31                              & 1.30                               \\ \hline
DocRED    & 30289        & 511                             & 2.67                               \\ \hline
PubMed-DU & 58761        & 15                              & 1.22                               \\ \hline
\end{tabular}
}
\caption{Datasets's statistics}
\label{table:dataset}
\end{table*}

\par
Based on the given training set, we consider that the model is aware of a general domain anatomy, i.e., the sets of possible entity types and relation types are known, and we would like to identify which of them are depicted in a given corpus. In both cases of \texttt{EntityPairOverlap} and \texttt{SingleEntityOverlap} type text snippets, there is one main entity type from which all the other entity types can be found by performing only one hop traversal in the general domain's metagraph. The class of \texttt{Normal} text snippets is a broader case in which one can identify heterogeneous connectivity patterns among the entity types represented. Yet, a sentence typically describes facts that are expected to be connected somehow, thus the entity types included in such texts usually are not more than 1 or 2 hops away from each other in the general metagraph. In the light of the considerations above, we propose to sort the relations in a breadth-first-search (BFS) order starting from a specific node (entity type) in the general metagraph. In this way, we confine the output in a much lower dimensional space by adhering to a semantically meaningful order.

\par
 Inspired by state-of-the-art approaches in the field of neural machine translation, our model architecture is a multi-layer bidirectional Transformer. We follow the lead of \citeauthor{attention} in implementing the architecture,  with the only difference that we adopt a learned positional encoding instead of a static one (see Appendix for further details on the positional encoding). As the overall architecture of the encoder and the decoder are otherwise the same as in \citeauthor{attention}, we omit an in-depth description of the Transfomer model and refer readers to original paper.

\par
To boost the model's performance, we also propose an ensemble approach exploiting different Transformers and  aggregating their results to construct the domain's metagraph. Each of the Transformers differs in the selected ordering of the \say{relation} vocabulary. The selection of different starting entity type for the breadth-first-search will lead to different orderings. We expect that multiple orderings could facilitate the prediction of different connection patterns that can not be easily detected using a single ordering. The sequence of steps for an ensemble domain understanding is the following: Firstly, train $k$ Transformers using different orderings. Secondly, given a set of text snippets, predict sequences of relations using all the Transformers. Finally, use late fusion to aggregate the results and form the final predictions.

It is worth mentioning that in the last step, we omit the underlying ordering that we follow in each model and we perform a relation-based aggregation.  We examine each relation separately in order to include it or not in the final metagraph. For the aggregation step, we use the standard Wisdom of Crowds (WOC) \cite{marbach2012wisdom} consensus technique, yet other consensus methods can also be leveraged for the task. The overall structure of our approach is summarized in Figure \ref{fig:arch}.

\section{Experiments}

\begin{table*}[t]
\centering
\begin{tabular}{|c|c|c|c|}
\hline
Dataset                    & Model                                        & Accuracy                     & F1 score                     \\ \hline  \hline
\multirow{5}{*}{WebNLG}    & CNN   \cite{nguyen2015relation}*             & 0.8156 $\pm$ 0.0071          & 0.9459 $\pm$ 0.0021          \\ \cline{2-4} 
                           & RNN   \cite{zhou2016attention}*              & 0.8517 $\pm$0.0058           & 0.9543 $\pm$ 0.0021          \\ \cline{2-4} 
                           & Transformer - unordered                      & \textbf{0.8798 $\pm$0.0053}  & \textbf{0.9646 $\pm$ 0.0018} \\ \cline{2-4} 
                           & Transformer - BFS$_{\textrm{record\_label}}$ & \textbf{0.9000 $\pm$ 0.0046} & \textbf{0.9699 $\pm$ 0.0013} \\ \cline{2-4} 
                           & Transformer - WOC k=20                       & \textbf{0.9235 $\pm$ 0.0014} & \textbf{0.9780 $\pm$ 0.0003} \\ \hline \hline
\multirow{5}{*}{NYT}       & CNN   \cite{nguyen2015relation}*             & 0.7341 $\pm$ 0.0035          & \textbf{0.8385 $\pm$ 0.0025} \\ \cline{2-4} 
                           & RNN    \cite{zhou2016attention}*             & \textbf{0.7520 $\pm$ 0.0027} & \textbf{0.8353 $\pm$ 0.0029} \\ \cline{2-4} 
                           & Transformer - unordered                      & 0.7426 $\pm$ 0.0061          & 0.8009 $\pm$ 0.0057          \\ \cline{2-4} 
                           & Transformer - BFS$_{\textrm{person}}$        & \textbf{0.7491 $\pm$ 0.0048} & 0.8049 $\pm$ 0.0073          \\ \cline{2-4} 
                           & Transformer - WOC k=8                        & \textbf{0.7669 $\pm$ 0.0011} & \textbf{0.8307 $\pm$ 0.0006} \\ \hline \hline
\multirow{5}{*}{DocRED}    & CNN    \cite{nguyen2015relation}*            & 0.1096 $\pm$ 0.0073          & 0.4434 $\pm$ 0.0133          \\ \cline{2-4} 
                           & RNN    \cite{zhou2016attention}*             & 0.2178$\pm$ 0.0088           & 0.6192 $\pm$ 0.0093          \\ \cline{2-4} 
                           & Transformer - unordered                      & \textbf{0.4869 $\pm$0.0069}  & \textbf{0.7081 $\pm$0.0032}  \\ \cline{2-4} 
                           & Transformer - BFS$_{\textrm{ORG}}$           & \textbf{0.5252 $\pm$ 0.0048} & \textbf{0.7133 $\pm$ 0.0049} \\ \cline{2-4} 
                           & Transformer - WOC k=6                        & \textbf{0.5722 $\pm$ 0.0001} & \textbf{0.7607 $\pm$ 0.0001} \\ \hline \hline
\multirow{5}{*}{PubMed-DU} & CNN    \cite{nguyen2015relation}*            & 0.5573 $\pm$ 0.0030          & \textbf{0.7063 $\pm$ 0.0048} \\ \cline{2-4} 
                           & RNN    \cite{zhou2016attention}*             & \textbf{0.5772 $\pm$ 0.0048} & \textbf{0.7234 $\pm$ 0.0032} \\ \cline{2-4} 
                           & Transformer - unordered                      & 0.5499 $\pm$ 0.0059          & 0.6725 $\pm$ 0.0056          \\ \cline{2-4} 
                           & Transformer - BFS$_{\textrm{Species}}$       & \textbf{0.5691 $\pm$ 0.0109} & 0.6752 $\pm$ 0.0045          \\ \cline{2-4} 
                           & Transformer - WOC k=5                        & \textbf{0.5946 $\pm$ 0.0001} & \textbf{0.7132 $\pm$ 0.0004} \\ \hline
\end{tabular}
\caption{Comparison of CNN model, RNN model and Transformer-based methods on WebNLG, NYT, DocRED and PubMed-DU datasets. *The architecture of the CNN and RNN models has been modified to exclude the component which provides information about the position of the entities in the text snippet.}
\label{table:results}

\end{table*}

\begin{table*}[]
\centering
\resizebox{0.9\textwidth}{!}{
\begin{tabular}{|c|c|c|c|c|c|}
\hline
Dataset                                                                              & Model                                        & \begin{tabular}[c]{@{}c@{}}Edges\\  F1-score\end{tabular} & \begin{tabular}[c]{@{}c@{}}Nodes\\  F1-score\end{tabular} & \begin{tabular}[c]{@{}c@{}}Degree JSD\end{tabular} & \begin{tabular}[c]{@{}c@{}}Eigenvector\\  JSD\end{tabular} \\ \hline  \hline
\multirow{5}{*}{WebNLG}                                                              & CNN     \cite{nguyen2015relation}*           & \textbf{0.9747}                                           & \textbf{0.9879}                                           & \textbf{0.1836}                                    & \textbf{0.2059}                                            \\ \cline{2-6} 
                                                                                     & RNN     \cite{zhou2016attention}*            & 0.9639                                                    & 0.9735                                                    & 0.2708                                             & 0.2364                                                     \\ \cline{2-6} 
                                                                                     & Transformer - unordered                      & 0.9598                                                    & \textbf{0.9775}                                           & 0.2380                                             & 0.2280                                                     \\ \cline{2-6} 
                                                                                     & Transformer - BFS$_{\textrm{record\_label}}$ & \textbf{0.9806}                                           & \textbf{0.9772}                                           & \textbf{0.1923}                                    & \textbf{0.1593}                                            \\ \cline{2-6} 
                                                                                     & Transformer - WOC k=5                        & \textbf{0.9808}                                           & \textbf{0.9772}                                           & \textbf{0.1765}                                    & \textbf{0.1261}                                            \\ \hline \hline
\multirow{5}{*}{NYT}                                                                 & CNN     \cite{nguyen2015relation}*           & \textbf{0.9059}                                           & \textbf{0.9800}                                           & 0.0564                                             & 0.0832                                                     \\ \cline{2-6} 
                                                                                     & RNN   \cite{zhou2016attention}*              & \textbf{0.9205}                                           & 1                                                         & \textbf{0}                                         & \textbf{0}                                                 \\ \cline{2-6} 
                                                                                     & Transformer - unordered                      & 0.8184                                                    & \textbf{0.9800}                                           & 0.0967                                             & 0.1396                                                     \\ \cline{2-6} 
                                                                                     & Transformer - BFS$_{\textrm{person}}$        & \textbf{0.8806}                                           & 0.9666                                                    & \textbf{0}                                         & \textbf{0}                                                 \\ \cline{2-6} 
                                                                                     & Transformer - WOC k=8                        & 0.8672                                                    & \textbf{1}                                                & \textbf{0}                                         & \textbf{0}                                                 \\ \hline \hline
\multirow{5}{*}{DocRED}                                                              & CNN      \cite{nguyen2015relation}*          & 0.4819                                                    & 0.9019                                                    & 0.5717                                             & 0.6965                                                     \\ \cline{2-6} 
                                                                                     & RNN    \cite{zhou2016attention}*             & 0.6823                                                    & 0.9714                                                    & 0.5187                                             & 0.6954                                                     \\ \cline{2-6} 
                                                                                     & Transformer - unordered                      & \textbf{0.7530}                                           & \textbf{1}                                                & \textbf{0.2950}                                    & \textbf{0.5997}                                            \\ \cline{2-6} 
                                                                                     & Transformer - BFS$_{\textrm{PER}}$           & \textbf{0.7830}                                           & \textbf{0.9777 }                                          & \textbf{0.2892}                                    & \textbf{0.4267}                                            \\ \cline{2-6} 
                                                                                     & Transformer - WOC k=6                        & \textbf{0.8045}                                           & \textbf{1}                                                & \textbf{0.2349}                                    & \textbf{0.3688}                                            \\ \hline \hline
\multirow{5}{*}{\begin{tabular}[c]{@{}c@{}}PubMed-DU\\ Covid-19 domain\end{tabular}} & CNN      \cite{nguyen2015relation}*          & 0.9140                                                    & \textbf{0.9888}                                           & 0.4175                                             & 0.4791                                                     \\ \cline{2-6} 
                                                                                     & RNN    \cite{zhou2016attention}*             & \textbf{0.9736}                                           & \textbf{0.9888}                                           & \textbf{0.1002}                                    & \textbf{0.1579}                                            \\ \cline{2-6} 
                                                                                     & Transformer - unordered                      & \textbf{0.9631}                                                    & \textbf{1}                                                & 0.3253                                             & \textbf{0.3987}                                                     \\ \cline{2-6} 
                                                                                     & Transformer - BFS$_{\textrm{Chemical}}$          & 0.9583                                                    & 0.9777                                                    & \textbf{0.2880}                                             & 0.4136                                            \\ \cline{2-6} 
                                                                                     & Transformer - WOC k=5                        & \textbf{0.9789}                                           & \textbf{0.9888}                                           & \textbf{0.2048}                                    & \textbf{0.2299}                                            \\ \hline
\end{tabular}
}

\caption{Evaluation of metagraph's reconstruction on the four datasets using CNN, RNN and Transformer-based models. For the PubMed-DU dataset, we focus only on the COVID-19 domain. *The architecture of the CNN and RNN models has been modified to exclude the component which provides information about the position of the entities in the text snippet.}
\label{table:graph_results}

\end{table*}

\par
We evaluate our Transformer-based approach against three baselines on a selection of datasets representing different domains. As baselines, we use CNN and RNN based methods influenced by \cite{nguyen2015relation} and \cite{zhou2016attention} respectively. For the CNN based method, we slightly modified the architecture to exclude the component which provides information about the position of the entities in the text snippet, as we do not have such information available in our task.
Additionally, we also include a Transformer-based model without applying any ordering in the target sequences as an extra baseline.
% We compare based on accuracy and F1-score. Accuracy is computed at an instance level as we examine how many target sentences are correct over all the testing set. The ordering of the target sentence is not assessed during the evaluation as we only examine the existence of each relation in the target and not its position. 

\par
To our knowledge, there is no standard dataset available for the relation type extraction task in the literature. However there is a plethora of published datasets for the standard task of relation extraction that can be utilized for our case with limited effort. For our task, the leveraged datasets should contain tuples of texts and their respective sets of relation types. We use WebNLG \cite{webnlg}, NYT \cite{riedel_nyt} and DocRED \cite{docred}, three of the most popular datasets for relation extraction. Both NYT and DocRED datasets provide the needed information such as entity types and relation type for the triplets of each instance. Thus their transformation for our task can be conducted by just converting these triplets to the relation type format, for instance the triplet (x,y,z) will be transformed as type(x).type(y).type(z). On the other hand, WebNLG doesn't share such information for the entity types and thus manual curation is needed. Therefore, all the possible entities are examined and replaced with the proper entity type. For the WebNLG dataset, we avoid including rare entity and relation types which are occurred less than 10 times in the dataset. We either omit them or replace them with similar or more general types that exists in it.

\par 
To emphasize the application of such model in the scientific document understanding, we produce a new task-specific dataset called PubMed-DU related to the general health domain. We download paper abstracts from PubMed focusing on work related to 4 specific health subdomains: Covid-19, mental health, breast cancer and coronary heart disease. We split the abstracts into sentences. The entities and their types for each text have been extracted using PubTator \cite{wei2019pubtator}. The available entity types are Gene, Mutation, Chemical, Disease and Species. The respective relation types are in form x.to.y where x and y are two of the possible entity types. We assume that the relations are symmetric. For text annotation, the following rule was used: a text has the relation x.to.y if two entities with types x and y co-occurred in the text and the syntax path between them contains at least one keyword of this relation type. These keywords have been manually identified based on the provided instances and are words, mainly verbs, related to the relation. Table \ref{table:dataset} depicts the statistics of all four utilised datasets.

\par
For all datasets, we use the same model parameters. Specifically, we use Adam \cite{kingma2014adam} optimizer with a learning rate of 0.0005. The gradients norm is clipped to 1.0 and dropout \cite{lecun2015deep} is set to 0.1. Both encoder and decoder consist of 2 layers with 10 attention heads each, the positional feed-forward hidden dimension is 512. Lastly, we  utilize the token embedding layers using GloVe  pretrained  word  embeddings \cite{pennington2014glove} which have dimensionality of m=300. Our code and the datasets are available at \url{https://github.com/christofid/DomainUnderstanding}.

\par
The evaluation of the models is performed both at instance and graph level. In the instance level, we examine the ability of the model to predict the relation types that exist in a given text. To investigate this, we use F1-score and accuracy. F1-score is the harmonic mean of model's precision and recall. Accuracy is computed at an instance level and it measures for how many of the testing texts, the model manage to infer correctly the whole set of their relation types. For the metagraph level evaluation, we use our model to predict the metagraph of a domain and we examine how close to the actual metagraph is. For this comparison, we utilize F1-score for both edges and nodes of the metagraph as well as the similarity of the distribution of the degree and eigenvector centrality \cite{zaki2014data} of the two metagraphs. For the comparison of the centralities distribution, we construct the histogram of the centralities for each graph using 10 fixed size bins and we utilize  Jensen-Shannon Divergence (JSD) metric \cite{JSD} to examine the similarity of the two distributions (see Appendix for the definition of JSD). We have selected degree and eigenvector centralities as the former gives as localized structure information as measure the importance of a node based on the direct connections of it and the latter gives as a broader structure information as measure the importance of a node based on infinite walks.

\subsection*{Instance level evaluation of the models}

\par
To study the performance of our model, we perform 10 independent runs each with different random splitting of the datasets into training, validation and testing set. Table \ref{table:results} depicts the median value and the standard error of the baselines and our method for the two metrics.
Our method is better in terms of accuracy for all the four datasets and in terms of F1-score for the WebNLG and DocRED datasets. For the NYT and PubMed-DU datasets, the F1-score of CNN and RNN models outperform our approach. We observed that the baseline models profit from the fact that, in these datasets, the majority of the instances depict only one relation and many of the relations appear in a limited number of instances. In general, there is lack of sequences of relations that hinders the Transformer's ability to learn the underlying distribution in these two cases(see Appendix). Lastly, the decreased performances of all the models in the DocRED dataset is due to the long tail characteristic that this dataset shows as 66\% of the relations appeared in no more than 50 instances (see Appendix).

\begin{figure*}[h]
\centering
         %\includesvg[scale=0.21]{images/kg_plot_n.svg}
        \includegraphics[scale=0.2]{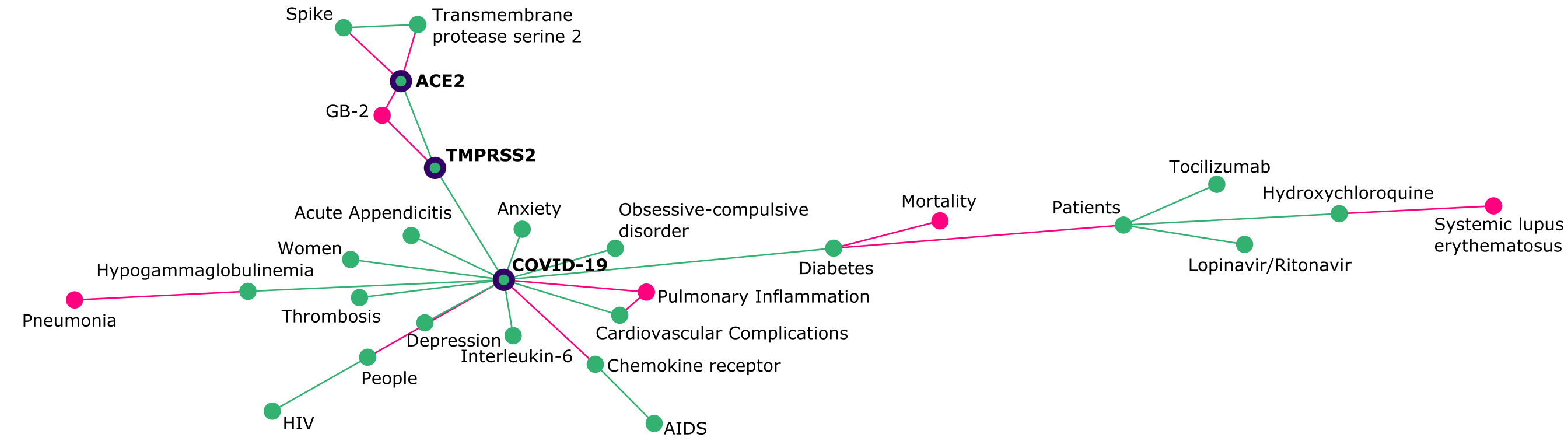}

\caption{KG  extracted from 24 text snippets related to the COVID-19 domain using our model and the respective attention analysis. Green color means that the respective node/edge exists in both actual and predicted graph, while pink color  means that this element exists in the actual but not in the predicted graph. Entities in bold indicate the path that has been extracted from the text \say{ace2 and tmprss2 variants and expression as candidates to sex and country differences in covid-19 severity in italy.}. }
        \label{fig:kg_meta}
\end{figure*}

\subsection*{Metagraph level evaluation of the models}

The above comparisons focus only on the ability of the model to predict the relation types given a text snippet. Since our ultimate goal  is to infer the domain's metagraph from a given corpus, we divide the testing sets of the datasets into small corpora and we attempt to define their domain using our model. For WebNLG, NYT and DocRED dataset, 10 artificial corpora and their respective domains have been created by selecting randomly 10 instances from each of the testing sets. We set two constraints into this selection to assure that the produced metagraphs are meaningful. Firstly, each subdomain should have a connected metagraph and secondly each existing relation type is appeared at least two times in the provided instances. For the PubMed-DU dataset, we already know the existence of 4 subdomains in it, so we focus on the inference of them. For each subdomain, we select randomly 100 instances from the testing set that belong to this subdomain and we attempt to produce the domain based on them. We infer the relation types for each instance and then we generate the domain's metagraph by including all the relation types  that were found in the instances. Then, we compare how close the actual domain's metagraph and the predicted metagraph are.

\par
Table \ref{table:graph_results} presents the results of the evaluation of the predicted versus the actual domain's metagraph for 10 subdomains extracted from the testing set of the WebNLG, NYT and DocRED datasets. All the presented values for these datasets are the mean over all the 10 subdomains. For the PubMed-DU dataset, we include only the Covid-19 subdomain case. Results for the remaining subdomains of this dataset can be found in the Appendix.  Our approach using Transformer + BFS based ordering outperforms or is close to the baselines for all cases in terms of edges and nodes F1-score. Furthermore, the degree and eigenvector centralities distribution of the generated metagraphs using our method are closer to the groundtruth in comparison to other methods in all cases. This indicates that the graphs produced with our method are both element-wise and structurally closer to the actual ones. More detailed comparisons of the different methods at both instance and metagraph level have been included in the Appendix.

\par
The ensemble variant of our approach, based on the WOC consensus strategy, outperforms the simple Transformer + BFS ordering in all cases. Based on the evaluation at both instance and metagraph level, our ensemble variant seems to be the most reliable approach for the task of domain's relation type extraction as it achieves some of the best scores for any dataset and metric. 

% \begin{figure}
%     \centering
%     \includegraphics[scale=0.21]{kg_metagraph.png}
%     \caption{Metagraph (left) and the KG (right) extracted from 12 text snippets related to the United States using our model and the respective attention analysis. The colors in the nodes/edges mean the following: green exists in both actual and predicted graphs, blue exists in the actual but not in the predicted graph, red exists in the predicted but not in the actual. }
%     \label{fig:kg_meta}
% \end{figure}

\subsection*{Towards automated KG generation}

The proposed domain understanding method enables the inference of the domain of interest and its components.
This enables a partial automation and a speed up of the KG generation process as, without manual intervention, we are able to identify the metagraph, and inherently the needed models for the entity and relation extraction in the context of the domain of interest.
To achieve this, we adopt a Transformer-based approach that relies heavily on attention mechanisms.
Recent efforts are focusing on the analysis of such attention mechanisms to explain and interpret the predictions and the quality of the models \cite{vig2019analyzing,hoover2019exbert}. Interestingly, it has been shown how the analysis of the attention pattern can elucidate complex relations between the entities fed as input to the Transformer, e.g., mapping atoms in chemical reactions with no supervision \cite{schwaller_hoover_reymond_strobelt_laino_2020}.
Even if it is out of the scope of our current work, we observe that a similar analysis of the attention patterns in our model can identify not only parts of text in which relations exist but directly the entities of the respective triplets.
To illustrate this and emphasize its application in the domain understanding field, we extract 24 text instances from the PubMed-DU dataset related to the COVID-19 domain. After generating the domain's metagraph, we analyze the attention to triples to build a KG. We rely on the syntax dependencies to propagate the attention weights throughout the connected tokens and we examine the noun chunks to extract the entities of interest based on their accumulated attention weight (see Appendix for further details). We select the head which achieves the best accuracy in order to generate the KG.
Figure \ref{fig:kg_meta} depicts the generated metagraph and the KG. Using the aforementioned attention analysis, we manage to achieve 82\% and 64\% accuracy in the entity extraction and the relation extraction respectively.
These values might not be able to compete the state of the art respective models and the investigation is limited in only few instances. Yet it indicates that a completely unsupervised generation based on attention analysis is possible and deserves further investigation.

\section{Conclusion}
\label{asdf}
Herein, we proposed a method to speed up the knowledge acquisition process of any domain specific KG application by defining the domain of interest in an automated manner. This is achieved by using a Transformer-based approach to estimate the metagraph representing the schema of the domain. Such schema  can indicate the proper and needed tools for the actual entity and relation extraction. Thus our method can be considering as the stepping stone in any KG generation pipeline. The evaluation and the comparison over different datasets against state-of-the-art methods indicates that our approach produces accurately the metagraph. Especially, in datasets where text instances contain multiple relation types our model outperforms the baselines. This is an important observation as text describing multiple relations is the most common scenario. Based on that and relying on the capability of the transformers to catch longer dependencies, future investigation of how our model performs in  larger pieces of texts, like full paragraphs, could be interesting and indicate a clearer advantage of our work. The needed definition of a general domain for the training phase might be a limitation of this method. However, schema and data from already existing KGs can be utilized for training purposes. Unsupervised or semi-supervised extension of this work can also be explored in the future to mitigate the issue. 

 Our work paves the way towards an automated knowledge acquisition, as our model minimizes the need of human intervention in the process. So in the near future the currently needed manual curation can be avoided and lead to faster and more accurate knowledge acquisition. Interestingly, using the PubMed-DU dataset, we underline that our method can be utilized for scientific documents. The inference of their domain can assist both in their general understanding but also lead to more robust knowledge acquisition from them. As a side effect, it is also important to notice that, such attention-based model can be directly applied to triplet extraction from the text without retraining and without supervision. Triplet extraction in an unsupervised way represents a breakthrough, especially if combined with most recent advances in zero-shot learning for NER \cite{li2020survey,pasupat2014zero,guerini2018toward}. Further analysis of our Transformer-based approach could give a better insight into these capabilities.

\bibliographystyle{aaai}
\bibliography{bibliography}

\appendix
\section{Appendix}
\subsection{Learned positional encoding }
\label{app:positional}
In our model, we adopt a learned positional encoding instead of a static one. Specifically, the tokens are passed through a standard embedding layer as a first step in the encoder. The model has no recurrent layers and therefore it has no idea about the order of the tokens within the sequence. To overcome this, we utilize a  second embedding layer called a positional embedding layer. This is a standard embedding layer where the input is not the token itself but the position of the token within the sequence, starting with the first token, the $<$sos$>$ (start of sequence) token, in position 0. The position embedding has a "vocabulary" size equal to the maximum length of the input sequence. The token embedding and positional embedding are element-wise summed together to get the final token embedding which contains information about both the token and its position within the sequence. This final token embedding is then provided as input in the stack of attention layers of the encoder.

\subsection{Dataset characteristics}
\label{app:docred}

\begin{figure*}
        \centering
        \begin{subfigure}[b]{0.475\textwidth}
            \centering
            \includegraphics[width=\textwidth]{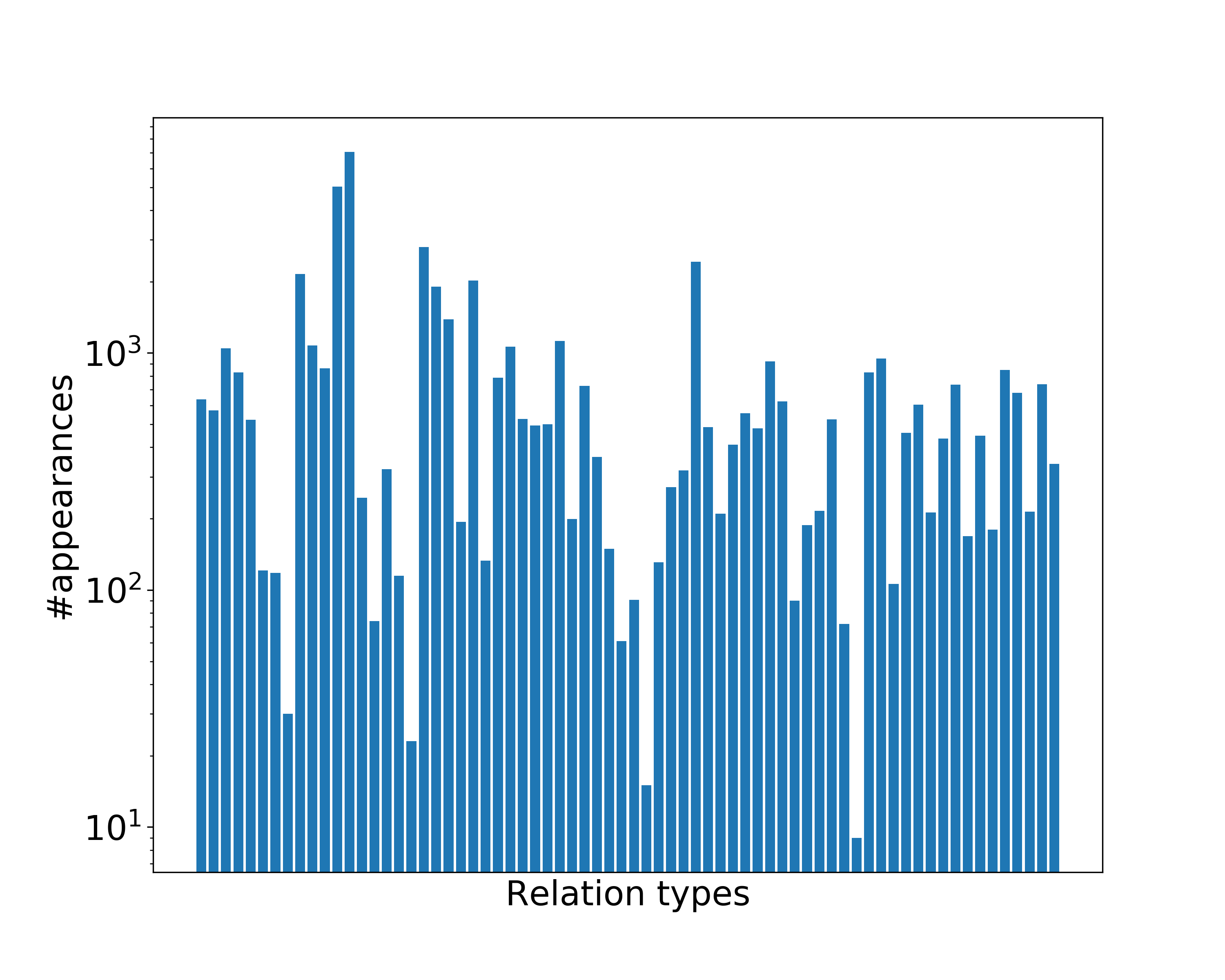}
            \caption[]%
            {{\small WebNLG}}    
            \label{fig:mean and std of net14}
        \end{subfigure}
        \hfill
        \begin{subfigure}[b]{0.475\textwidth}  
            \centering 
            \includegraphics[width=\textwidth]{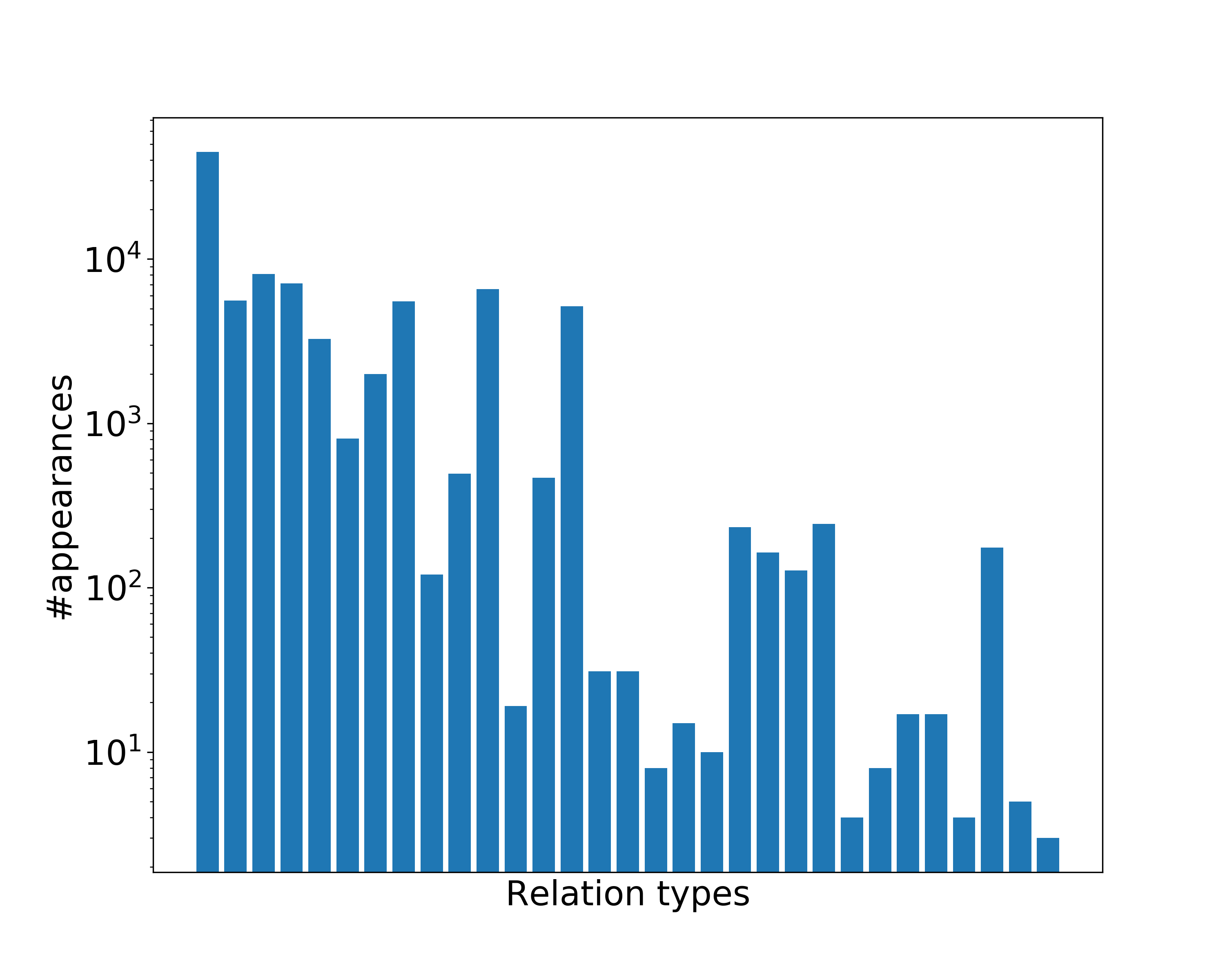}
            \caption[]%
            {{\small NYT}}    
            \label{fig:mean and std of net24}
        \end{subfigure}
        \vskip\baselineskip
        \begin{subfigure}[b]{0.475\textwidth}   
            \centering 
            \includegraphics[width=\textwidth]{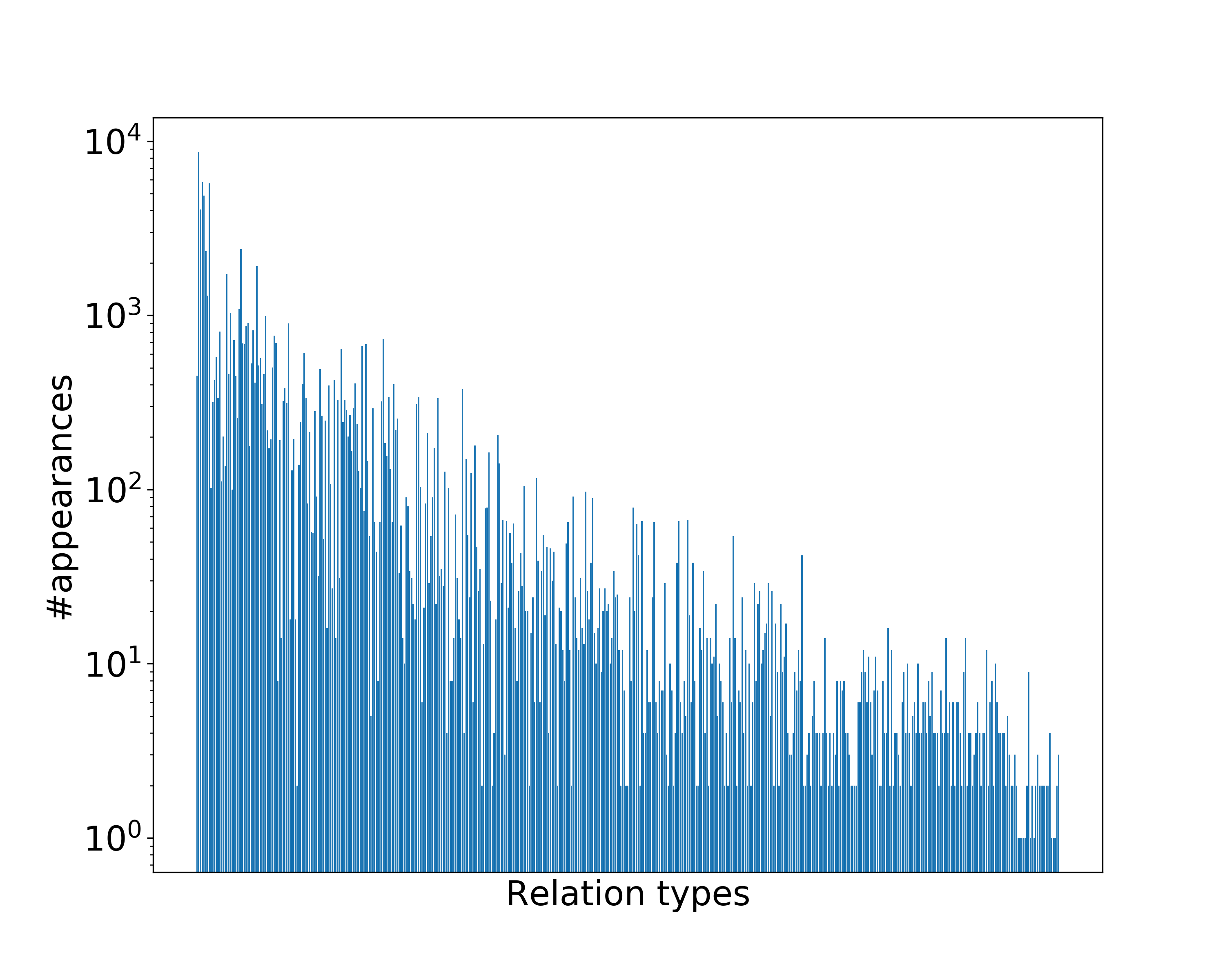}
            \caption[]%
            {{\small DocRED}}    
            \label{fig:mean and std of net34}
        \end{subfigure}
        \hfill
        \begin{subfigure}[b]{0.475\textwidth}   
            \centering 
            \includegraphics[width=\textwidth]{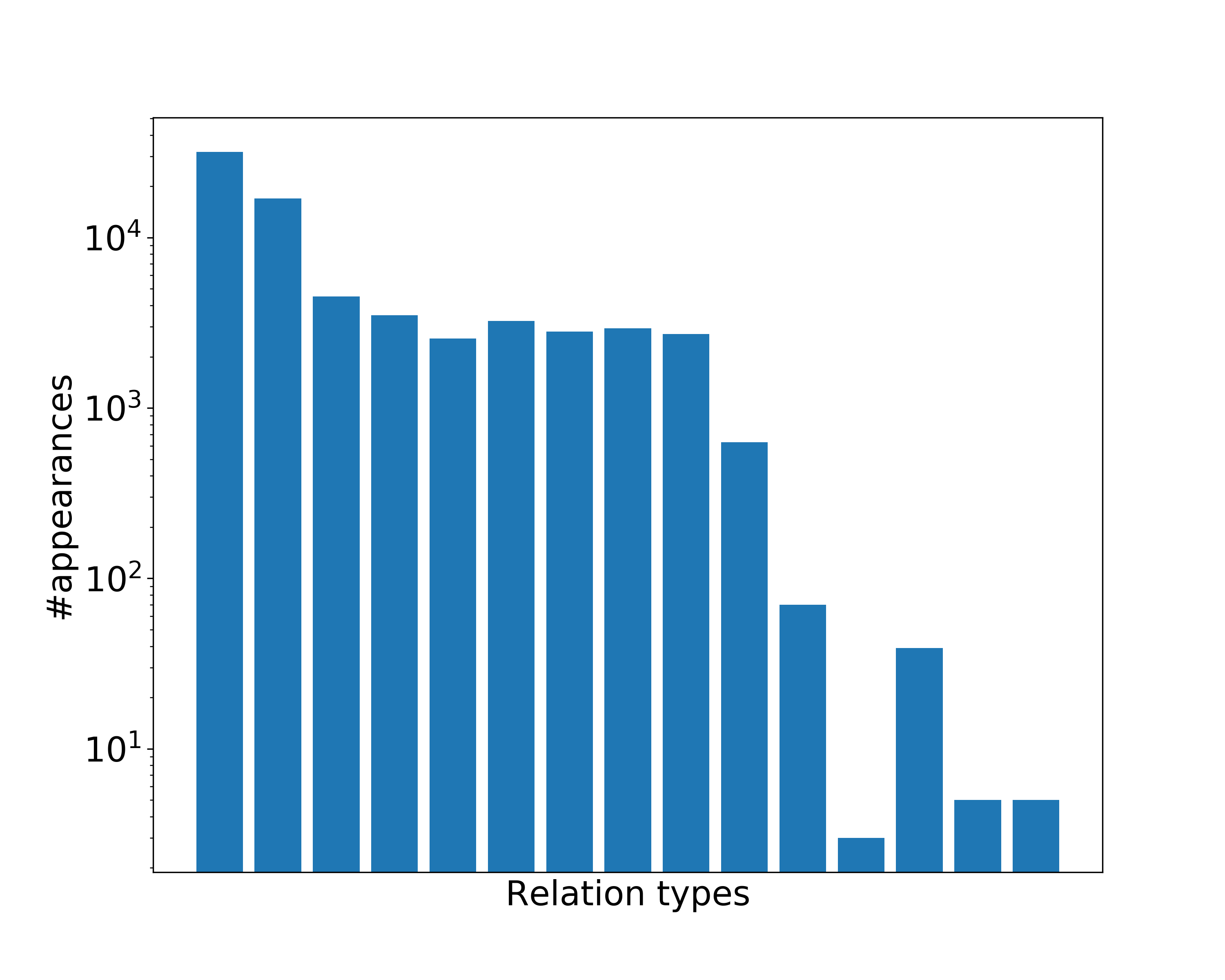}
            \caption[]%
            {{\small PubMed-DU}}    
            \label{fig:mean and std of net44}
        \end{subfigure}
        \caption
        {\small Appearances distribution for the relation types of all utilized datasets.} 
        \label{fig:rel_dist}
    \end{figure*}

For a better understanding of the datasets, we analyzed the distribution of occurrences for all relation types. These distributions are depicted in Figure \ref{fig:rel_dist}. A percentage of relation types with number of appearances close or less to 10 is observed for all datasets. The lack of many examples can pose problems in the learning process for these specific relation types. This is highlighted especially in the DocRED case, as we attributed the decreased performances of all the models in this dataset in its long tail characteristic that it holds. Especially for the DocRED, almost the 50\% of the relations appeared in no more than 10 instances and the 66\% of the relations appeared in no more than 50 instances (\ref{fig:rel_dist}).

\subsection{Jensen-Shannon distance}
\label{app:jensen}
The Jensen-Shannon divergence metric between two probability vectors p and q is defined as:

$$ \sqrt{ \frac{D(p \parallel m)+D(q \parallel m)} {2}} $$

where m is the pointwise mean of p and q  and D is the Kullback-Leibler divergence.

The Kullback-Leibler divergence for two probability vectors p and q of length n is defined as:

$$ D(p \parallel q) = \sum_{i=1}^{n}p_ilog_2(\frac{p_i}{q_i}) $$

The Jensen–Shannon metric is bounded by 1, given that we use the base 2 logarithm.

\subsection{Triplets extraction based on attention analysis}
\label{app:attention}

In this section, the procedure of automated triplets extraction based on the predicted relation types and the respective attention weights is described . We generate an undirected graph that connects the tokens of the sentence based on their syntax dependencies for each instance. Then for each different predicted relation type, we define the final attention weights of a token based on the attention weights of itself and its neighbors in the syntax dependencies graph. Let $a^r$ be the attention vector of a predefined model's attention head, which contains all the attention weights related to the relation type $r$. The final attention weight $w$ of the token $i$ for the relation $r$ is defined as:

  $$ w^r_i = 2*a^r_i + \sum_{j \in neig_i}{a^r_j} $$

where $neig_i$ is the set containing all the neighbors of $i$ in the syntax dependencies graph. Then for each noun chunk $k$ ($n_k$) of the text we compute its total attention weight for the relation type $r$ as:

$$ n_k^r  = \sum_{j \in nc_k} f(w^r_j) $$

where $nc_k$ is the set of tokens which belong to the $n_k$ and $f$ is a function defined as \[ f(w^r_j) =
  \begin{cases}
    w^r_j       & \quad \text{if } j \text{ is  stop-word}\\
    2*w^r_j  & \quad \text{if } j \text{ is not  stop-word}
  \end{cases}
\]

Finally, we extract as entities which are connected via the relation type $r$ the two noun chunks with the highest  weight $n^r$. As this work is a proof of concept rather than an actual method, the selection of the attention head is based on whichever gives as the best outcome. Yet, in actual scenarios it is recommended the use of a training set, based on which the optimal head will be identified. For the creation of the syntax dependencies graph and the extraction of the noun chunks of the text we use spacy\footnote{https://spacy.io/} and its  \texttt{en\_core\_web\_lg} pretrained model. Table \ref{app:att} includes all the texts that have used in the proof of concept that is presented in the main paper and the respective predicted triplets for each of them.

\begin{table*}[]
\centering
\resizebox{\textwidth}{!}{
\begin{tabular}{|l|l|}
\hline
Text                                                                                                                                                                                                                                                                                        & Predicted triplets                                                                                        \\ \hline
acute deep vein thrombosis in covid-19 hospitalized patients.                                                                                                                                                                                                                               & (thrombosis, Disease.to.Disease, COVID 19)                                                                   \\ \hline
\begin{tabular}[c]{@{}l@{}}(ace-2) receptor for its attachment similar to sars-cov-1, which\\ is followed by priming of spike protein by transmembrane protease serine 2 \\ (tmprss2) which can be targeted by a proven inhibitor of tmprss2, camostat.\end{tabular}                        & (spike, Gene.to.Gene, Transmembrane protease serine 2)                                                       \\ \hline
\begin{tabular}[c]{@{}l@{}}temporal trends in decompensated heart failure and outcomes during covid-19:\\ a multisite report from heart failure referral centres in london.\end{tabular}                                                                                                    & (heart failure, Disease.to.Disease, COVID-19)                                                                \\ \hline
\begin{tabular}[c]{@{}l@{}}prevalence, risk factors and clinical correlates of depression in quarantined\\ population during the covid-19 outbreak.\end{tabular}                                                                                                                            & (depression, Disease.to.Disease, COVID-19)                                                                   \\ \hline
tocilizumab plus glucocorticoids in severe and critically covid-19 patients.                                                                                                                                                                                                                & (Tocilizumab, Chemical.to.Species, patients)                                                                 \\ \hline
\begin{tabular}[c]{@{}l@{}}effects of progressive muscle relaxation on anxiety and sleep quality in \\ patients with covid-19.\end{tabular}                                                                                                                                                 & (anxiety, Disease.to.Disease, COVID-19)                                                                      \\ \hline
special section: covid-19 among people living with hiv.                                                                                                                                                                                                                                     & (people, Disease.to.Species, HIV)                                                                            \\ \hline
\begin{tabular}[c]{@{}l@{}}ace2 and tmprss2 variants and expression as candidates to sex and country\\ differences in covid-19 severity in italy.\end{tabular}                                                                                                                              & \begin{tabular}[c]{@{}l@{}}(ACE2, Gene.to.Gene, TMPRSS2)\\ (COVID-19, Disease.to.Gene, TMPRSS2)\end{tabular} \\ \hline
did the covid-19 pandemic cause a delay in the diagnosis of acute appendicitis?                                                                                                                                                                                                             & (COVID-19, Disease.to.Disease, Acute Appendicitis)                                                           \\ \hline
\begin{tabular}[c]{@{}l@{}}practice observed in managing gynaecological problems in post-menopausal \\ women during the covid-19 pandemic.\end{tabular}                                                                                                                                     & (COVID-19, Disease.to.Species, women)                                                                        \\ \hline
\begin{tabular}[c]{@{}l@{}}hypogammaglobulinemia causing pneumonia: an overlooked curable entity \\ in the chaotic covid-19 pandemic.\end{tabular}                                                                                                                                          & (hypogammaglobulinemia, Disease.to.Disease, COVID-19)                                                        \\ \hline
\begin{tabular}[c]{@{}l@{}}chemokine receptor gene polymorphisms and covid-19: could knowledge\\ gained from hiv/aids be important?\end{tabular}                                                                                                                                            & (AIDS, Disease.to.Gene, Chemokine receptor)                                                                  \\ \hline
serotonin syndrome in two covid-19 patients treated with lopinavir/ritonavir.                                                                                                                                                                                                               & (lopinavir/ritonavir, Chemical.to.Species, patients)                                                         \\ \hline
\begin{tabular}[c]{@{}l@{}}mortality rate and predictors of mortality in hospitalized covid-19 \\ patients with diabetes.\end{tabular}                                                                                                                                                      & (COVID-19, Disease.to.Disease, Diabetes)                                                                     \\ \hline
\begin{tabular}[c]{@{}l@{}}in conclusion, self-reported depression occurred at an early stage in convalescent\\ covid-19 patients, and changes in immune function were apparent during\\ short-term follow-up of these patients after discharge.\end{tabular}                               & (depression, Disease.to.Disease, COVID-19)                                                                   \\ \hline
gb-2 inhibits ace2 and tmprss2 expression: in vivo and in vitro studies.                                                                                                                                                                                                                    & (ACE2, Gene.to.Gene, TMPRSS2)                                                                                \\ \hline
preemptive interleukin-6 blockade in patients with covid-19.                                                                                                                                                                                                                                & (COVID-19, Disease.to.Gene, interleukin-6)                                                                   \\ \hline
\begin{tabular}[c]{@{}l@{}}response to: 'clinical course of covid-19 in patients with systemic lupus\\ erythematosus under long-term treatment with hydroxychloroquine' by carbillon et al.\end{tabular}                                                                                    & (hydroxychloroquine, Chemical.to.Species, patients)                                                          \\ \hline
obsessive-compulsive disorder during the covid-19 pandemic                                                                                                                                                                                                                                  & (Obsessive-compulsive disorder, Disease.to.Disease, COVID-19)                                                \\ \hline
\begin{tabular}[c]{@{}l@{}}repeated monitoring of ferritin, interleukin-6, c-reactive protein, lactic acid dehydrogenase, \\ and erythrocyte sedimentation rate during covid-19 treatment may assist the prediction of\\ disease severity and evaluation of treatment effects.\end{tabular} & (COVID-19, Disease.to.Gene, interleukin-6)                                                                   \\ \hline
\begin{tabular}[c]{@{}l@{}}respiratory and pulmonary complications in head and neck cancer patients: evidence-based \\ review for the covid-19 era.\end{tabular}                                                                                                                            & (head and neck cancer, Disease.to.Disease, COVID-19)                                                         \\ \hline
\begin{tabular}[c]{@{}l@{}}targeting the immune system for pulmonary inflammation and cardiovascular complications \\ in covid-19 patients.\end{tabular}                                                                                                                                    & (Cardiovascular Complications, Disease.to.Disease, COVID-19)                                                 \\ \hline
risk of peripheral arterial thrombosis in covid-19.                                                                                                                                                                                                                                         & (Thrombosis, Disease.to.Disease, COVID-19)                                                                   \\ \hline
\begin{tabular}[c]{@{}l@{}}preadmission diabetes-specific risk factors for mortality in hospitalized patients with \\ diabetes and coronavirus disease 2019.\end{tabular}                                                                                                                   & (Diabetes, Disease.to.Disease, COVID-19)                                                                     \\ \hline
\end{tabular}
}
\caption{Utilized texts from PubMed-DU dataset and their respective extracted triplets based on attention analysis.}
\label{app:att}
\end{table*}

\subsection{Models's comparison}
\label{app:results_detail}
Table \ref{table:results_app} presents a detailed evaluation of our approach and the baselines models. We have included the 3 best BFS ordering variants (in terms of accuracy) and 3 consensus variants. To cover the range of all the available values of k, [1, number of entity types], we select a case with just a few Transformers, one with a value close to half of the total number of entity types and one close to the total number of entity types. For each different $k$ value, we utilize the top k best orderings based on their accuracy. In addition to the per instance accuracy and the per relation F1-score, the table also includes the per relation precision and the recall of each model. Our proposed method, especially its ensemble variant, produces the best outcome in all datasets apart from the NYT case where the CNN and RNN models manage to be more precise. This is attributed to the characteristics of NYT dataset, where there are many one-only relation instances.

\begin{table*}
\centering
\resizebox{\textwidth}{!}{
\begin{tabular}{|c|c|c|c|c|c|}
\hline
Dataset                 & Model                                        & Accuracy                     & Precision                    & Recall                       & F1 score                     \\ \hline  \hline
\multirow{9}{*}{WebNLG} & CNN  \cite{nguyen2015relation}*              & 0.8156 $\pm$ 0.0071          & 0.9550 $\pm$ 0.0029          & 0.9370 $\pm$ 0.0049          & 0.9459 $\pm$ 0.0021          \\ \cline{2-6} 
                        & RNN  \cite{zhou2016attention}*               & 0.8517 $\pm$0.0058           & 0.9614 $\pm$0.0022           & 0.9472$\pm$0.0043            & 0.9543 $\pm$0.0021           \\ \cline{2-6} 
                        & Transformer - unordered                      & 0.8798 $\pm$0.0053           & 0.9678 $\pm$0.0032           & 0.9614 $\pm$0.0042           & 0.9646 $\pm$0.0018           \\ \cline{2-6} 
                        & Transformer - BFS$_{\textrm{occupation}}$    & 0.8987$\pm$0.0068            & 0.9693$\pm$0.0030            & 0.9705 $\pm$ 0.0035          & 0.9699 $\pm$ 0.0018          \\ \cline{2-6} 
                        & Transformer - BFS$_{\textrm{music\_genre}}$  & 0.8983 $\pm$ 0.0053          & 0.9694 $\pm$0.0039           & 0.9703 $\pm$0.0030           & 0.9699 $\pm$ 0.0015          \\ \cline{2-6} 
                        & Transformer - BFS$_{\textrm{record\_label}}$ & 0.9000 $\pm$ 0.0046          & 0.9707 $\pm$ 0.0031          & 0.9691 $\pm$0.0028           & 0.9699 $\pm$0.0013           \\ \cline{2-6} 
                        & Transformer - WOC k=5                        & \textbf{0.9210 $\pm$ 0.0017} & \textbf{0.9789 $\pm$0.0016}  & \textbf{0.9755 $\pm$ 0.0013} & \textbf{0.9772 $\pm$ 0.0004} \\ \cline{2-6} 
                        & Transformer - WOC k=20                       & \textbf{0.9235 $\pm$ 0.0014} & \textbf{0.9786 $\pm$0.0006}  & \textbf{0.9774 $\pm$ 0.0004} & \textbf{0.9780 $\pm$0.0003}  \\ \cline{2-6} 
                        & Transformer - WOC k=45                       & \textbf{0.9235 $\pm$ 0.0002} & \textbf{0.9795 $\pm$0.0001}  & \textbf{0.9767 $\pm$0.0001}  & \textbf{0.9781 $\pm$0.0001}  \\ \hline \hline
\multirow{9}{*}{NYT}    & CNN    \cite{nguyen2015relation}*            & 0.7341 $\pm$ 0.0035          & \textbf{0.8873 $\pm$ 0.0032} & 0.7948 $\pm$ 0.0053          & \textbf{0.8385 $\pm$ 0.0025} \\ \cline{2-6} 
                        & RNN    \cite{zhou2016attention}*             & 0.7520 $\pm$ 0.0027          & \textbf{0.8530 $\pm$0.0041}  & 0.8183 $\pm$0.0047           & \textbf{0.8353 $\pm$0.0029}  \\ \cline{2-6} 
                        & Transformer - unordered                      & 0.7426 $\pm$ 0.0061          & 0.8059 $\pm$0.0079           & 0.7960 $\pm$ 0.0070          & 0.8009 $\pm$0.0057           \\ \cline{2-6} 
                        & Transformer - BFS$_{\textrm{location}}$      & 0.7461 $\pm$ 0.0053          & 0.8093 $\pm$0.0067           & 0.7988 $\pm$0.0067           & 0.8040 $\pm$0.0043           \\ \cline{2-6} 
                        & Transformer - BFS$_{\textrm{person}}$        & 0.7491 $\pm$0.0048           & 0.8119 $\pm$ 0.0036          & 0.8022 $\pm$0.0032           & 0.8049 $\pm$ 0.0073          \\ \cline{2-6} 
                        & Transformer - BFS$_{\textrm{company}}$       & 0.7461 $\pm$0.0081           & 0.8056 $\pm$0.0088           & 0.8043 $\pm$0.0117           & 0.8049 $\pm$0.0073           \\ \cline{2-6} 
                        & Transformer - WOC k=4                        & \textbf{0.7547 $\pm$0.0038}  & 0.8162 $\pm$ 0.0047          & \textbf{0.8355 $\pm$ 0.0022} & 0.8257 $\pm$0.0023           \\ \cline{2-6} 
                        & Transformer - WOC k=8                        & \textbf{0.7669 $\pm$0.0011}  & \textbf{0.8325 $\pm$ 0.0025} & \textbf{0.8289 $\pm$0.0018}  & 0.8307 $\pm$ 0.0006          \\ \cline{2-6} 
                        & Transformer - WOC k=12                       & \textbf{0.7698 $\pm$ 0.0007} & 0.8381 $\pm$0.0017           & \textbf{0.8296 $\pm$0.0009}  & \textbf{0.8320 $\pm$0.0004}  \\ \hline \hline
\multirow{9}{*}{DocRED} & CNN     \cite{nguyen2015relation}*           & 0.1096 $\pm$ 0.0073          & 0.7838 $\pm$ 0.0081          & 0.3094 $\pm$ 0.0140          & 0.4434 $\pm$ 0.0133          \\ \cline{2-6} 
                        & RNN     \cite{zhou2016attention}*            & 0.2178$\pm$ 0.0088           & 0.7716 $\pm$0.0100           & 0.5173 $\pm$0.0143           & 0.6192 $\pm$ 0.0093          \\ \cline{2-6} 
                        & Transformer - unordered                      & 0.4869 $\pm$0.0069           & 0.7365 $\pm$0.0156           & 0.6822 $\pm$0.0127           & 0.7081 $\pm$0.0032           \\ \cline{2-6} 
                        & Transformer - BFS$_{\textrm{LOC}}$           & 0.5235 $\pm$ 0.0049          & 0.7145 $\pm$ 0.0112          & 0.7053 $\pm$0.0076           & 0.7098 $\pm$ 0.0040          \\ \cline{2-6} 
                        & Transformer - BFS$_{\textrm{PER}}$           & 0.5234 $\pm$ 0.0077          & 0.7234 $\pm$ 0.0179          & 0.7029 $\pm$ 0.0091          & 0.7128 $\pm$0.0060           \\ \cline{2-6} 
                        & Transformer - BFS$_{\textrm{ORG}}$           & 0.5252 $\pm$ 0.0048          & 0.7216 $\pm$ 0.0068          & 0.7053 $\pm$ 0.0069          & 0.7133 $\pm$ 0.0049          \\ \cline{2-6} 
                        & Transfomer - WOC k=4                         & \textbf{0.5678 $\pm$0.0037}  & \textbf{0.7939 $\pm$0.0137}  & \textbf{0.7227 $\pm$0.0103}  & \textbf{0.7564 $\pm$ 0.0032} \\ \cline{2-6} 
                        & Transformer - WOC k=5                        & \textbf{0.5697 $\pm$ 0.0016} & \textbf{0.8012 $\pm$0.0112}  & \textbf{0.7210 $\pm$0.0053}  & \textbf{0.7589 $\pm$0.0025}  \\ \cline{2-6} 
                        & Transformer - WOC k=6                        & \textbf{0.5722 $\pm$ 0.0001} & \textbf{0.7970 $\pm$ 0.0035} & \textbf{0.7276 $\pm$ 0.0022} & \textbf{0.7607 $\pm$ 0.0001} \\ \hline \hline
\multirow{9}{*}{PubMed-DU} & CNN \cite{nguyen2015relation}*               & 0.5573 $\pm$ 0.0030          & \textbf{0.7856 $\pm$ 0.0063} & 0.6417 $\pm$ 0.0100          & 0.7063 $\pm$ 0.0048          \\ \cline{2-6} 
                        & RNN \cite{zhou2016attention}*                & \textbf{0.5772 $\pm$ 0.0041} & \textbf{0.7401 $\pm$ 0.0051} & \textbf{0.7075 $\pm$ 0.0051}         & \textbf{0.7234 $\pm$ 0.0032} \\ \cline{2-6} 
                        & Transformer - unordered                      & 0.5693 $\pm$ 0.0075          & 0.6920 $\pm$ 0.0099          & 0.6837 $\pm$ 0.0111          & 0.6877 $\pm$ 0.0047          \\ \cline{2-6} 
                        & Transformer - BFS$_{\textrm{Species}}$       & 0.5707 $\pm$ 0.0060          & 0.6878 $\pm$ 0.0084          & 0.6921 $\pm$ 0.0097          & 0.6898 $\pm$ 0.0040          \\ \cline{2-6} 
                        & Transformer - BFS$_{\textrm{Chemical}}$      & 0.5703 $\pm$ 0.0048          & 0.6864 $\pm$ 0.0056          & 0.6942 $\pm$ 0.0057          & 0.6903 $\pm$ 0.0048          \\ \cline{2-6} 
                        & Transformer - BFS$_{\textrm{Gene}}$          & 0.5718 $\pm$ 0.0068          & 0.6872 $\pm$ 0.0058          & 0.6906 $\pm$ 0.0110          & 0.6889 $\pm$ 0.0065          \\ \cline{2-6} 
                        & Transformer - WOC k=3                        & \textbf{0.5909 $\pm$ 0.0070} & 0.7261 $\pm$ 0.0162          & \textbf{0.6958 $\pm$ 0.0031} & 0.7106 $\pm$ 0.0094          \\ \cline{2-6} 
                        & Transformer - WOC k=4                        & 0.5685 $\pm$ 0.0069          & 0.6864 $\pm$ 0.0109          & \textbf{0.7418 $\pm$ 0.0033} & \textbf{0.7130 $\pm$ 0.0074} \\ \cline{2-6} 
                        & Transformer - WOC k=5                        & \textbf{0.5960 $\pm$ 0.0001} & \textbf{0.7346 $\pm$ 0.0028} & 0.6974 $\pm$ 0.0025 & \textbf{0.7155 $\pm$ 0.0004} \\ \hline
\end{tabular}
}
\caption{Comparison of CNN, RNN and Transformer-based methods on WebNLG, NYT, DocRED and PubMed-DU datasets for the relation type extractiont task. *The architecture of the CNN and RNN models has been modified to exclude the component which provides information about the position of the entities in the text snippet.}
\label{table:results_app}
\end{table*}

Similarly, in tables \ref{table:graph_results_app} and  \ref{table:graph_results_pubmed_app} we perform an in-depth metagraph level evaluation of the models. For all three datasets, our proposed method and its ensemble extension produce the best or one of the top-3 best outcomes.

\begin{table*}

\centering
\resizebox{\textwidth}{!}{
\begin{tabular}{|c|c|c|c|c|c|}
\hline
\multicolumn{1}{|l|}{Dataset}                 & Model                           & \begin{tabular}[c]{@{}c@{}}Edges\\  F1-score\end{tabular} & \begin{tabular}[c]{@{}c@{}}Nodes\\  F1-score\end{tabular} & \begin{tabular}[c]{@{}c@{}}Degree JSD\end{tabular} & \begin{tabular}[c]{@{}c@{}}Eigenvector\\  JSD\end{tabular} \\ \hline \hline
\multicolumn{1}{|l|}{\multirow{9}{*}{WebNLG}} & CNN      \cite{nguyen2015relation}*                       & 0.9747                                                    & \textbf{0.9879}                                           & 0.1836                                                       & 0.2059                                                            \\ \cline{2-6}
\multicolumn{1}{|l|}{}                        & RNN    \cite{zhou2016attention}*                         & 0.9639                                                    & 0.9735                                                    & 0.2708                                                       & 0.2364                                                            \\ \cline{2-6}
\multicolumn{1}{|l|}{}                        & Transformer - unordered         & 0.9598                                                    & 0.9775                                                    & 0.2380                                                       & 0.2280                                                            \\ \cline{2-6}
\multicolumn{1}{|l|}{}                        & Transformer - BFS$_{\textrm{occupation}}$    & \textbf{0.9831}                                           & 0.9805                                                    & 0.1743                                                       & 0.1840                                                            \\ \cline{2-6}
\multicolumn{1}{|l|}{}                        & Transformer - BFS$_{\textrm{music\_genre}}$   & 0.9755                                                    & 0.9746                                                    & \textbf{0.1131}                                              & 0.1306                                                            \\ \cline{2-6}
\multicolumn{1}{|l|}{}                        & Transformer - BFS$_{\textrm{record\_label}}$  & 0.9806                                                    & 0.9772                                                    & 0.1923                                                       & 0.1593                                                            \\ \cline{2-6}
\multicolumn{1}{|l|}{}                        & Transformer - WOC k=5           & 0.9808                                                    & 0.9772                                                    & 0.1765                                                       & \textbf{0.1261}                                                   \\ \cline{2-6}
\multicolumn{1}{|l|}{}                        & Transformer - WOC k=20          & \textbf{0.9864}                                           & \textbf{0.9840}                                           & \textbf{0.1456}                                              & \textbf{0.070}                                                    \\ \cline{2-6}
\multicolumn{1}{|l|}{}                        & Transformer - WOC k=45          & \textbf{0.9930}                                           & \textbf{0.9916}                                           & \textbf{0.1313}                                              & \textbf{0.0891}                                                   \\ \hline \hline
\multirow{9}{*}{NYT}                          & CNN      \cite{nguyen2015relation}*                       & \textbf{0.9059}                                           & 0.9800                                                    & 0.0564                                                       & 0.0832                                                            \\ \cline{2-6}
                                              & RNN     \cite{zhou2016attention}*                        & \textbf{0.9205}                                           & \textbf{1}                                                & \textbf{0}                                                   & \textbf{0}                                                        \\ \cline{2-6}
                                              & Transformer - unordered         & 0.8184                                                    & 0.9800                                                    & 0.0967                                                       & 0.1396                                                            \\ \cline{2-6}
                                              & Transformer - BFS$_{\textrm{location}}$       & 0.8141                                                    & 0.9800                                                    & 0.0832                                                       & 0.0832                                                            \\ \cline{2-6}
                                              & Transformer - BFS$_{\textrm{person}}$         & \textbf{0.8806}                                           & 0.9666                                                    & \textbf{0}                                                   & \textbf{0}                                                        \\ \cline{2-6}
                                              & Transformer - BFS$_{\textrm{company}}$        & 0.8305                                                    & 0.9657                                                    & \textbf{0}                                                   & \textbf{0}                                                        \\ \cline{2-6}
                                              & Transformer - WOC k=4           & 0.8442                                                    & \textbf{1}                                                & \textbf{0}                                                   & \textbf{0}                                                        \\ \cline{2-6}
                                              & Transformer - WOC k=8           & 0.8672                                                    & \textbf{1}                                                & \textbf{0}                                                   & \textbf{0}                                                        \\ \cline{2-6}
                                              & Transformer - WOC k=12          & 0.8666                                                    & \textbf{1}                                                & \textbf{0}                                                   & \textbf{0}                                                        \\ \hline \hline
\multirow{9}{*}{DocRED}                       & CNN      \cite{nguyen2015relation}*                       & 0.4819                                                    & 0.9019                                                    & 0.5717                                                       & 0.6965                                                            \\ \cline{2-6}
                                              & RNN      \cite{zhou2016attention}*                       & 0.6823                                                    & 0.9714                                                    & 0.5187                                                       & 0.6954                                                            \\ \cline{2-6}
                                              & Transformer - unordered         & 0.7530                                                    & \textbf{1}                                                & 0.2950                                                       & 0.5997                                                            \\ \cline{2-6}
                                              & Transformer - BFS$_{\textrm{LOC}}$            & 0.7710                                                    & \textbf{1}                                                & 0.2744                                                       & 0.5140                                                            \\ \cline{2-6}
                                              & Transformer - BFS$_{\textrm{PER}}$            & 0.7830                                                    & 0.9777                                                    & 0.2892                                                       & \textbf{0.4267}                                                   \\ \cline{2-6}
                                              & Transformer - BFS$_{\textrm{ORG}}$            & 0.7243                                                    & 0.9777                                                    & 0.2892                                                       & \textbf{0.4267}                                                   \\ \cline{2-6}
                                              & Transformer - WOC k=4           & \textbf{0.7997}                                           & \textbf{1}                                                & \textbf{0.2714}                                              & 0.4787                                                            \\ \cline{2-6}
                                              & Transformer - WOC k=5           & \textbf{0.8090}                                           & \textbf{1}                                                & \textbf{0.2673}                                              & 0.4433                                                            \\ \cline{2-6}
                                              & Transformer - WOC k=6           & \textbf{0.8045}                                           & \textbf{1}                                                & \textbf{0.2349}                                              & \textbf{0.3688}                                                   \\ \hline
\end{tabular}
}
\caption{Evaluation of metagraph's reconstruction on WebNLG, NYT and DocRED datasets using CNN, RNN and Transformer-based models. *The architecture of the CNN and RNN models has been modified to exclude the component which provides information about the position of the entities in the text snippet.}
\label{table:graph_results_app}
\end{table*}

\begin{table*}[]
\resizebox{\textwidth}{!}{
\begin{tabular}{|c|c|c|c|c|c|}
\hline
Domain                                  & Model                                   & \begin{tabular}[c]{@{}c@{}}Edges\\  F1-score\end{tabular} & \begin{tabular}[c]{@{}c@{}}Nodes\\  F1-score\end{tabular} & \begin{tabular}[c]{@{}c@{}}Degree JSD\end{tabular} & \begin{tabular}[c]{@{}c@{}}Eigenvector\\  JSD\end{tabular} \\ \hline  \hline
\multirow{9}{*}{Covid-19}                & CNN      \cite{nguyen2015relation}*     & 0.9140                                                    & \textbf{0.9888}                                           & 0.4175                                             & 0.4791                                                     \\ \cline{2-6} 
                                         & RNN    \cite{zhou2016attention}*        & \textbf{0.9736}                                           & \textbf{0.9888}                                           & \textbf{0.1002}                                    & \textbf{0.1579}                                            \\ \cline{2-6} 
                                         & Transformer - unordered                 & 0.9631                                                    & \textbf{1}                                                & 0.3253                                             & 0.3987                                                     \\ \cline{2-6} 
                                         & Transformer - BFS$_{\textrm{Species}}$  & 0.9583                                                    & 0.9777                                                    & \textbf{0.2880}                                    & 0.4136                                                     \\ \cline{2-6} 
                                         & Transformer - BFS$_{\textrm{Chemical}}$ & 0.9583                                                    & 0.9777                                                    & \textbf{0.2880}                                    & 0.4136                                                     \\ \cline{2-6} 
                                         & Transformer - BFS$_{\textrm{Gene}}$     & 0.9525                                                    & 0.9777                                                    & 0.3344                                             & \textbf{0.2919}                                            \\ \cline{2-6} 
                                         & Transformer - WOC k=3                   & 0.9531                                                    & 0.9777                                                    & 0.4271                                             & 0.4096                                                     \\ \cline{2-6} 
                                         & Transformer - WOC k=4                   & \textbf{0.9642}                                           & 0.9777                                                    & 0.4271                                             & 0.4096                                                     \\ \cline{2-6} 
                                         & Transformer - WOC k=5                   & \textbf{0.9789}                                           & \textbf{0.9888}                                           & \textbf{0.2048}                                    & \textbf{0.2299}                                            \\ \hline \hline
\multirow{9}{*}{Breast cancer}           & CNN      \cite{nguyen2015relation}*     & 0.9135                                                    & \textbf{1}                                                & \textbf{0.3575}                                    & \textbf{0.3987}                                            \\ \cline{2-6} 
                                         & RNN     \cite{zhou2016attention}*       & \textbf{0.9730}                                           & \textbf{1}                                                & \textbf{0.1962}                                    & \textbf{0.2594}                                            \\ \cline{2-6} 
                                         & Transformer - unordered                 & 0.9367                                                    & 0.9777                                                    & 0.4534                                             & 0.4879                                                     \\ \cline{2-6} 
                                         & Transformer - BFS$_{\textrm{Species}}$  & \textbf{0.9531}                                           & 0.9777                                                    & 0.4453                                             & 0.4879                                                     \\ \cline{2-6} 
                                         & Transformer - BFS$_{\textrm{Chemical}}$ & \textbf{0.9531}                                           & 0.9777                                                    & 0.4453                                             & 0.4879                                                     \\ \cline{2-6} 
                                         & Transformer - BFS$_{\textrm{Gene}}$     & 0.9379                                                    & 0.9666                                                    & 0.5385                                             & 0.5343                                                     \\ \cline{2-6} 
                                         & Transformer - WOC k=3                   & 0.9525                                                    & 0.9777                                                    & 0.4089                                             & \textbf{0.4068}                                            \\ \cline{2-6} 
                                         & Transformer - WOC k=4                   & \textbf{0.9584}                                           & 0.9777                                                    & 0.3999                                             & 0.4414                                                     \\ \cline{2-6} 
                                         & Transformer - WOC k=5                   & 0.9514                                                    & \textbf{0.9888}                                           & \textbf{0.3821}                                    & \textbf{0.3987}                                            \\ \hline \hline
\multirow{9}{*}{Coronary heart diseases} & CNN      \cite{nguyen2015relation}*     & 0.8902                                                    & \textbf{0.9777}                                           & 0.4349                                             & 0.5865                                                     \\ \cline{2-6} 
                                         & RNN      \cite{zhou2016attention}*      & 0.9498                                                    & 0.9666                                                    & 0.3661                                             & 0.3391                                                     \\ \cline{2-6} 
                                         & Transformer - unordered                 & 0.9484                                                    & \textbf{0.9777}                                           & 0.4025                                             & 0.5153                                                     \\ \cline{2-6} 
                                         & Transformer - BFS$_{\textrm{Species}}$  & \textbf{0.9703}                                           & \textbf{0.9777}                                           & \textbf{0.2586}                                    & \textbf{0.2558}                                            \\ \cline{2-6} 
                                         & Transformer - BFS$_{\textrm{Chemical}}$ & \textbf{0.9703}                                           & \textbf{0.9777}                                           & \textbf{0.2586}                                    & \textbf{0.2558}                                            \\ \cline{2-6} 
                                         & Transformer - BFS$_{\textrm{Gene}}$     & \textbf{0.9756}                                           & \textbf{0.9777}                                           & \textbf{0.1659}                                    & \textbf{0.1571}                                            \\ \cline{2-6} 
                                         & Transformer - WOC k=3                   & 0.9644                                                    & \textbf{0.9777}                                           & 0.2705                                             & 0.3023                                                     \\ \cline{2-6} 
                                         & Transformer - WOC k=4                   & \textbf{0.9703}                                           & \textbf{0.9777}                                           & 0.2705                                             & 0.3023                                                     \\ \cline{2-6} 
                                         & Transformer - WOC k=5                   & 0.9644                                                    & \textbf{0.9777}                                           & 0.2705                                             & 0.3023                                                     \\ \hline \hline
\multirow{9}{*}{Mental health}           & CNN   \cite{nguyen2015relation}*                                  & 0.9419                                                    & \textbf{1}                                                & 0.2477                                             & 0.4634                                                     \\ \cline{2-6} 
                                         & RNN  \cite{zhou2016attention}*                                   & \textbf{0.9924}                                           & \textbf{1}                                                & \textbf{0.067}                                     & \textbf{0.0663}                                            \\ \cline{2-6} 
                                         & Transformer - unordered                 & 0.9697                                                    & \textbf{1}                                                & 0.2028                                             & 0.2601                                                     \\ \cline{2-6} 
                                         & Transformer - BFS$_{\textrm{Species}}$  & 0.9765                                                    & \textbf{1}                                                & 0.1503                                             & 0.3568                                                     \\ \cline{2-6} 
                                         & Transformer - BFS$_{\textrm{Chemical}}$ & 0.9765                                                    & \textbf{1}                                                & 0.1503                                             & 0.3568                                                     \\ \cline{2-6} 
                                         & Transformer - BFS$_{\textrm{Gene}}$     & \textbf{0.9849}                                           & \textbf{1}                                                & \textbf{0.1333}                                    & \textbf{0.1852}                                            \\ \cline{2-6} 
                                         & Transformer - WOC k=3                   & 0.9765                                                    & \textbf{1}                                                & 0.1503                                             & 0.3042                                                     \\ \cline{2-6} 
                                         & Transformer - WOC k=4                   & 0.9765                                                    & \textbf{1}                                                & 0.1503                                             & 0.3568                                                     \\ \cline{2-6} 
                                         & Transformer - WOC k=5                   & \textbf{0.9840}                                           & \textbf{1}                                                & \textbf{0.0840}                                    & \textbf{0.2378}                                            \\ \hline
\end{tabular}
}
\caption{Evaluation of metagraph's reconstruction on the 4 predefined subdomains of PubMed-DU dataset using CNN, RNN and Transformer-based models. *The architecture of the CNN and RNN models has been modified to exclude the component which provides information about the position of the entities in the text snippet.}
\label{table:graph_results_pubmed_app}

\end{table*}

\end{document}